\relax
\documentclass[letterpaper]{article} 
\usepackage{aaai20}  
\usepackage{multirow}
\usepackage{booktabs}
\usepackage{float}
\usepackage{pifont}
\usepackage{times}  
\usepackage{helvet} 
\usepackage{courier}  
\usepackage[hyphens]{url}  
\usepackage{graphicx} 
\usepackage{graphicx}  
\title{CBNet: A Novel Composite Backbone Network Architecture for Object Detection }

\author{Yudong Liu,\textsuperscript{1}  
	Yongtao Wang,\textsuperscript{1} Siwei Wang,\textsuperscript{1} TingTing Liang,\textsuperscript{1} \\ \Large\textbf{Qijie Zhao,\textsuperscript{1} Zhi Tang,\textsuperscript{1} Haibin Ling \textsuperscript{2}}\\  
\large 
\textsuperscript{1} Wangxuan Institute of Computer Technology, Peking University\\
\textsuperscript{2} Department of Computer Science , Stony Brook University \\
\{bahuangliuhe,wyt,wangsiwei,liangtingting,zhaoqijie,tangzhi\}@pku.edu.cn\\
haibin.ling@stonybrook.edu
}

\begin{document}

\maketitle

\begin{abstract}
	In existing CNN based detectors, the backbone network is a very important component for basic feature\footnote{Here and after, “basic feature” refers in particular to the features which are extracted by the backbone network and used as the input to other functional modules in the detector like detection head, RPN and FPN.} extraction, and the performance of the detectors highly depends on it. In this paper, we aim to achieve better detection performance by building a more powerful backbone from existing backbones like ResNet and ResNeXt. Specifically, we propose a novel strategy for assembling multiple identical backbones by composite connections between the adjacent backbones, to form a more powerful backbone named \emph{Composite Backbone Network} (CBNet). In this way, CBNet iteratively feeds the output features of the previous backbone, namely high-level features, as part of input features to the succeeding backbone, in a stage-by-stage fashion, and finally the feature maps of the last backbone (named Lead Backbone) are used for object detection. We show that CBNet can be very easily integrated into most state-of-the-art detectors and significantly improve their performances. For example, it boosts the mAP of FPN, Mask R-CNN and Cascade R-CNN on the COCO dataset by about 1.5 to 3.0 percent. Meanwhile, experimental results show that the instance segmentation results can also be improved. Specially, by simply integrating the proposed CBNet into the baseline detector Cascade Mask R-CNN, we achieve a new state-of-the-art result on COCO dataset (mAP of 53.3) with single model, which demonstrates great effectiveness of the proposed CBNet architecture. Code will be made available on https://github.com/PKUbahuangliuhe/CBNet.

\end{abstract}

\section{Introduction}
Object detection is one of the most fundamental problems in computer vision, which can serve a wide range of applications such as autonomous driving, intelligent video surveillance, remote sensing, and so on. In recent years, great progresses have been made for object detection thanks to the booming development of the deep convolutional networks \cite{krizhevsky2012imagenet}, and a few excellent detectors have been proposed, \textit{e.g.}, SSD  \cite{liu2015ssd}, Faster R-CNN \cite{ren2015faster}, Retinanet\cite{lin2018focal}, FPN \cite{lin2017feature}, Mask R-CNN \cite{he2017mask}, Cascade R-CNN \cite{cai18cascadercnn}, \textit{etc}.

\begin{figure}[t]
	\centering
	\includegraphics[width=\linewidth,height=.77\linewidth]{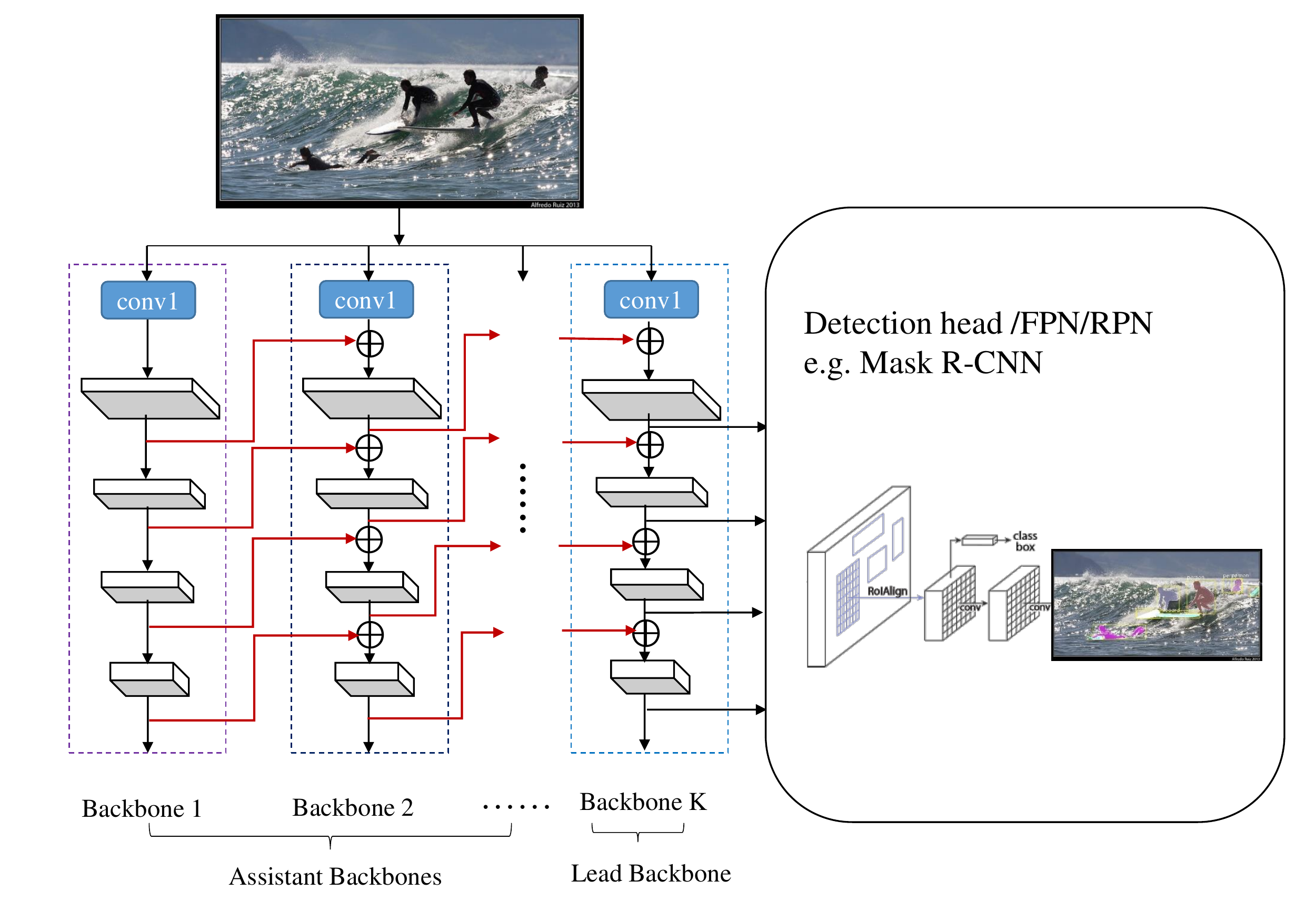}
	\caption{Illustration of the proposed Composite Backbone Network (CBNet) architecture for object detection. CBNet assembles multiple identical backbones (Assistant Backbones and Lead Backbone) by composite connections between the parallel stages of the adjacent backbones. In this way, CBNet iteratively feeds the output features of the previous backbone as part of input features to the succeeding backbone, in a stage-by-stage fashion, and finally outputs the features of the last backbone namely Lead Backbone for object detection. The red arrows represent composite connections.}
	\label{fig:1}
\end{figure}

\begin{table}[H]
	\centering 
	\begin{tabular}{l|ccc} 
		\toprule
		Backbone Network& $AP_{box}$ & $AP_{50}$ & $AP_{75}$ \\ 
		\hline
		\textit{ResNet50}& 41.5 & 60.1 & 45.5\\
		\textit{ResNet101}& 43.3 & 61.7 & 47.2\\
		\textit{ResNeXt101}& 45.9 & 64.4 & 50.2 \\
		\textit{ResNeXt152}& 48.3 & 67.0 & 52.8 \\
		\hline
		\textbf{\textit{Dual-ResNeXt152 (ours)}}& \textbf{50.0} & \textbf{68.8} & \textbf{54.6} \\
		\textbf{\textit{Triple-ResNeXt152 (ours)}}& \textbf{50.7} & \textbf{69.8} & \textbf{55.5}\\
		\bottomrule 
	\end{tabular}
	\caption{Results of the state-of-the-art detector Cascade Mask R-CNN on the COCO \texttt{test-dev} dataset \cite{lin2014microsoft} with different existing backbones and the proposed Composite Backbone Networks (Dual-ResNeXt152 and Triple-ResNeXt152), which are reproduced by Detectron \cite{Detectron2018,cai18cascadercnn}. It shows that deeper and larger backbones bring better detection performance, while our Composite Backbone Network architecture can further strengthen the existing very powerful backbones for object detection such as ResNeXt152.} 
	\label{table:backbone with cascade-mask rcnn} 
\end{table}

Generally speaking, in a typical CNN based object detector, a backbone network is used to extract basic features for detecting objects, which is usually designed for the image classification task and pretrained on the ImageNet dataset \cite{deng2009imagenet}. Not surprisingly, if a backbone can extract more representational features, its host detector will perform better accordingly. In other words, a more powerful backbone can bring better detection performance, as demonstrated in Table \ref{table:backbone with cascade-mask rcnn}. Hence, starting from AlexNet \cite{krizhevsky2012imagenet}, deeper and larger (\textit{i.e.}, more powerful) backbones have been exploited by the state-of-the-art detectors, such as VGG \cite{simonyan2014very}, ResNet \cite{he2016deep}, DenseNet    \cite{huang2017densely}, ResNeXt \cite{xie2017aggregated}. Despite encouraging results achieved by the state-of-the-art detectors based on deep and large backbones, there is still plenty of room for performance improvement. Moreover, it is very expensive to achieve better detection performance by designing a novel more powerful backbone and pre-training it on ImageNet. In addition, since almost all of the existing backbone networks are originally designed for image classification task, directly employing them to extract basic features for object detection may result in suboptimal performance.

To deal with the issues mentioned above, as illustrated in Figure ~\ref{fig:1}, we propose to assemble multiple identical backbones, in a novel way, to build a more powerful backbone for object detection. In particular, the assembled backbones are treated as a whole which we call Composite Backbone Network (CBNet). More specifically, CBNet consists of multiple identical backbones (specially called Assistant Backbones and Lead Backbone) and composite connections between neighbor backbones. From left to right, the output of each stage in an Assistant Backbone, namely higher-level features, flows to the parallel stage of the succeeding backbone as part of inputs through composite connections. Finally, the feature maps of the last backbone named Lead Backbone are used for object detection.  Obviously, the features extracted by CBNet for object detection fuse the high-level and low-level features of multiple backbones, hence improve the detection performance. It is worth mentioning that, \textit{we do not need to pretrain CBNet for training a detector integrated with it}. For instead, \textit{we only need to initialize each assembled backbone of CBNet with the pretrained model of the single backbone} which is widely and freely available today, such as ResNet and ResNeXt. In other words, adopting the proposed CBNet is more economical and efficient than designing a novel more powerful backbone and pre-training it on ImageNet.

On the widely tested MS-COCO benchmark~\cite{lin2014microsoft}, we conduct experiments by applying the proposed Composite Backbone Network to several state-of-the-art object detectors, such as FPN~\cite{lin2017feature}, Mask R-CNN~\cite{he2017mask} and Cascade R-CNN~\cite{cai18cascadercnn}. Experimental results show that the mAPs of all the detectors consistently increase by 1.5 to 3.0 percent, which demonstrates the effectiveness of our Composite Backbone Network. Moreover, with our Composite Backbone Network, the results of instance segmentation are also improved. Specially, using Triple-ResNeXt152, \textit{i.e.}, Composite Backbone Network architecture of three ResNeXt152 ~\cite{xie2017aggregated} backbones, we achieve the new state-of-the-art result on COCO dataset, that is, mAP of 53.3, outperforming all the published object detectors. 

To summarize, the major contributions of this work are two-fold:
\begin{itemize}
	\item We propose a novel method to build a more powerful backbone for object detection by assembling multiple identical backbones, which can significantly improve the performances of various state-of-the-art detectors.
	\item We achieve the new state-of-the-art result on the MSCOCO dataset with single model, that is, the mAP of 53.3 for object detection.
\end{itemize}

In the rest of the paper, after reviewing related work in Sec. \ref{section 2}, we describe in details the proposed CBNet for object detection in Sec. \ref{section 3}. Then, we report the experimental validation in Sec. \ref{section 4}. Finally, we draw the conclusion in Sec. \ref{section 6}.

\section{Related work}
\label{section 2}

\textbf{Object detection}
Object detection is a fundamental problem in computer vision. The state-of-the-art methods for general object detection can be briefly categorized into two major branches. The first branch contains one-stage methods such as YOLO \cite{Redmon_2016_CVPR}, SSD  \cite{liu2015ssd}, Retinanet \cite{lin2017focal}, FSAF \cite{zhu2019feature} and NAS-FPN \cite{ghiasi2019fpn}. The other branch contains two-stage methods such as Faster R-CNN \cite{ren2015faster}, FPN \cite{lin2017feature}, Mask R-CNN\cite{he2017mask}, Cascade R-CNN\cite{cai18cascadercnn} and Libra R-CNN \cite{pang2019libra}. Although breakthrough has been made and encouraging results have been achieved by the recent CNN based detectors, there is still large room for performance improvement. For example, on MS COCO benchmark \cite{lin2014microsoft}, the best publicly reported mAP is only 52.5 \cite{peng2018megdet}, which is achieved by model ensemble of four detectors.

\textbf{Backbone for Object detection}
Backbone is a very important component of a CNN based detector to extract basic features for object detection. Following the original works (\textit{e.g.}, R-CNN \cite{girshick2014rich} and OverFeat \cite{sermanet2013overfeat}) of applying deep learning to object detection, almost all of the recent detectors adopt the \textit{pretraining and fine-tuning} paradigm, that is, directly use the networks which are pre-trained for ImageNet classification task as their backbones. For instance, VGG \cite{simonyan2014very}, ResNet \cite{he2016deep}, ResNeXt \cite{xie2017aggregated} are widely used by the state-of-the-art detectors. Since these backbone networks are originally designed for image classification task, directly employing them to extract basic features for object detection may result in suboptimal performance. More recently, two sophisticatedly designed backbones, \textit{i.e.}, DetNet \cite{li2018detnet} and FishNet \cite{sun2018fishnet}, are proposed for object detection. These two backbones are specifically designed for the object detection task, and they still need to be pretrained for ImageNet classification task before training (fine tuning) the detector based on them. It is well known that designing and pretraining a novel and powerful backbone like them requires much manpower and computation cost. In an alternative way, we propose a more economic and efficient solution to build a more powerful backbone for object detection, by assembling multiple identical existing backbones (\textit{e.g.}, ResNet and ResNeXt).

\begin{figure}[t]
	\centering
	\includegraphics[scale=0.4]{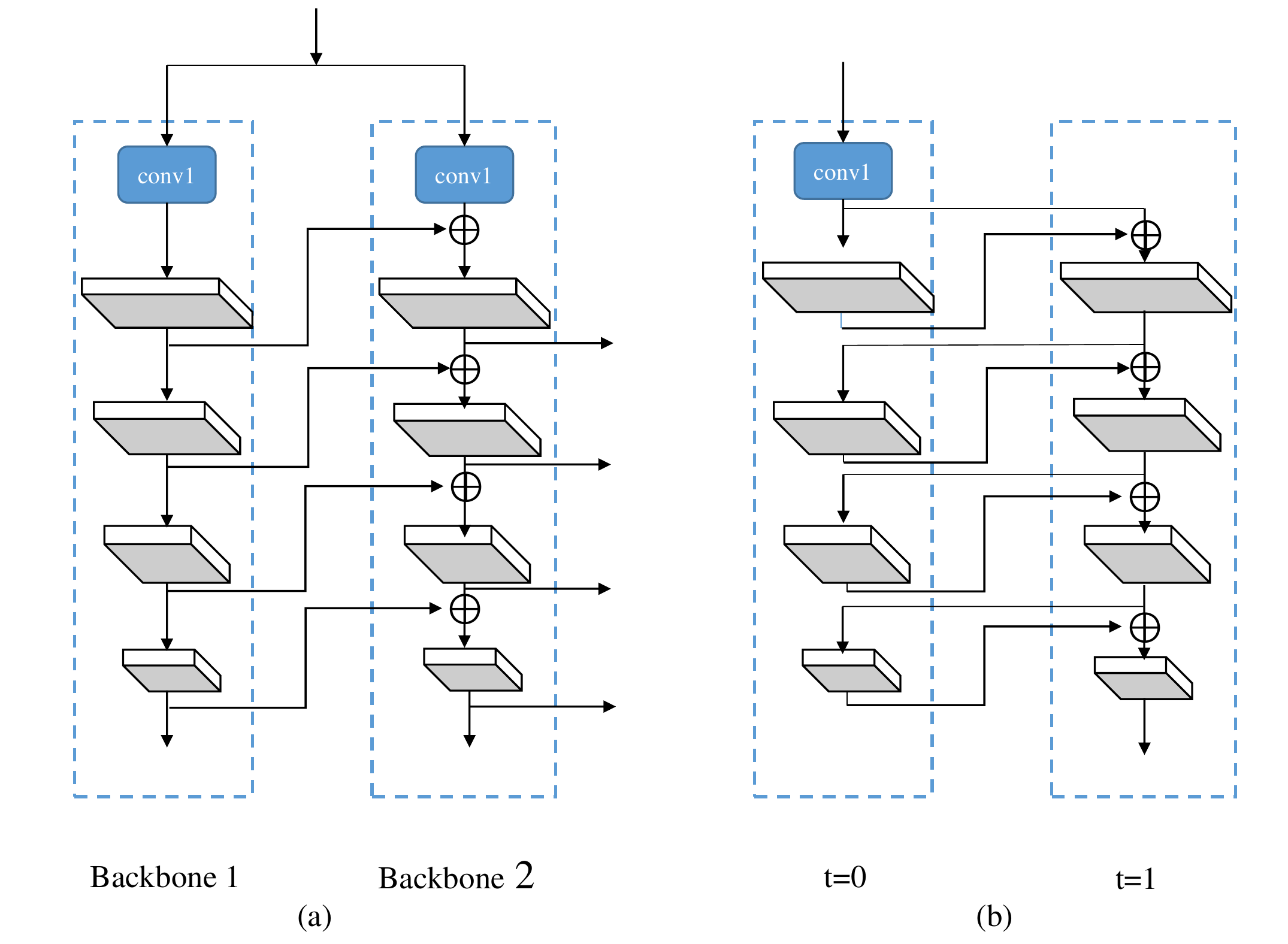}
	\caption{Comparison between (a). our proposed CBNet architecture ($K=2$) and  (b). the unrolled architecture of RCNN \cite{liang2015recurrent}($T=2$).}
	\label{fig:RCNN}
\end{figure}

\textbf{Recurrent Convolution Neural Network}
As shown in Figure ~\ref{fig:RCNN}, the proposed architecture of Composite Backbone Network is somewhat similar to an unfolded recurrent convolutional neural network (RCNN) \cite{liang2015recurrent} architecture. However, the proposed CBNet is quite different from this network. First, as illustrated in Figure ~\ref{fig:RCNN}, the architecture of CBNet is actually quite different, especially for the connections between the parallel stages. Second, in RCNN, the parallel stages of different time steps share the parameters, while in the proposed CBNet, the parallel stages of backbones do not share the parameters. Moreover, if we use RCNN as the backbone of a detector, we need to pretrain it on ImageNet. However, when we use CBNet, we do not need to pretrain it.


\section{Proposed method}
\label{section 3}
This section elaborates the proposed CBNet in detail. We first describe its architecture and variants in Section \ref{sec:arc} and Section \ref{sec:opcab} respectively. And then, we describe the structure of detection network with CBNet in Section \ref{sec:det}.

\subsection{Architecture of CBNet}
\label{sec:arc}

The architecture of the proposed CBNet consists of $K$ identical backbones ($K\ge2$). Specially, we call the case of K = 2 (as shown in Figure \ref{fig:RCNN}.a) as \textbf{Dual-Backbone (DB)} for simplicity, and the case of K=3 as \textbf{Triple- Backbone (TB)}.

As illustrated in Figure \ref{fig:1}, the CBNet architecture consists of two types of backbones: the Lead Backbone $B_{K}$ and the Assistant Backbones $B_{1}, B_{2}, ... , B_{K-1}$. Each backbone comprises $L$ stages (generally $L=5$), and each stage consists of several convolutional layers with feature maps of the same size. The $l$-th stage of the backbone implements a non-linear transformation $F^l(\cdot)$.

In the traditional convolutional network with only one backbone, the $l$-th stage takes the output (denoted as $x^{l-1}$) of the previous $l-1$-th stage as input, which can be expressed as:
\begin{equation}
\label{equa1}
x^l = F^l(x^{l-1}),  l\ge2.
\end{equation}
Unlike this, in the CBNet architecture, we novelly employ Assistant Backbones $B_{1}, B_{2}, ... , B_{k-1}$ to enhance the features of the Lead Backbone $B_k$, by iteratively feeding the output features of the previous backbone as part of input features to the succeeding backbone, in a stage-by-stage fashion. To be more specific, the input of the $l$-th stage of the backbone $B_k$ is the fusion of the output of the previous $l-1$-th stage of $B_k$ (denoted as $x_k^{l-1}$) and the output of the parallel stage of the previous backbone $B_{k-1}$ (denoted as $x_{k-1}^l$). This operation can be formulated as following:

\begin{equation}
\label{equa2}
x_k^l = F_k^l(x_k^{l-1} + g(x_{k-1}^l)),  l\ge2,
\end{equation}
where $g(\cdot)$ denotes the composite connection, which consists of a 1$\times$1 convolutional layer and batch normalization layer to reduce the channels and an upsample operation. As a result, the output features of the $l$-th stage in $B_{k-1}$ is transformed to the input of the same stage in $B_k$, and added to the original input feature maps to go through the corresponding layers. Considering that this composition style feeds the output of the adjacent higher-level stage of the previous backbone to the succeeding backbone, we call it as Adjacent Higher-Level Composition (AHLC).

%

For object detection task, only the output of Lead Backbone $x_K^{l}(l = 2,3,...L)$ are taken as the input of RPN/detection head, while the output of each stage of Assistant Backbones is forwarded into its adjacent backbone. Moreover, the $B_{1}, B_{2}, ... , B_{K-1}$ in CBNet can adopt various backbone architectures, such as ~\cite{he2016deep} or ResNeXt ~\cite{xie2017aggregated}, and can be initialized from the pre-trained model of the single backbone directly.



\begin{figure*}[t]
	\centering
	\includegraphics[scale=0.8]{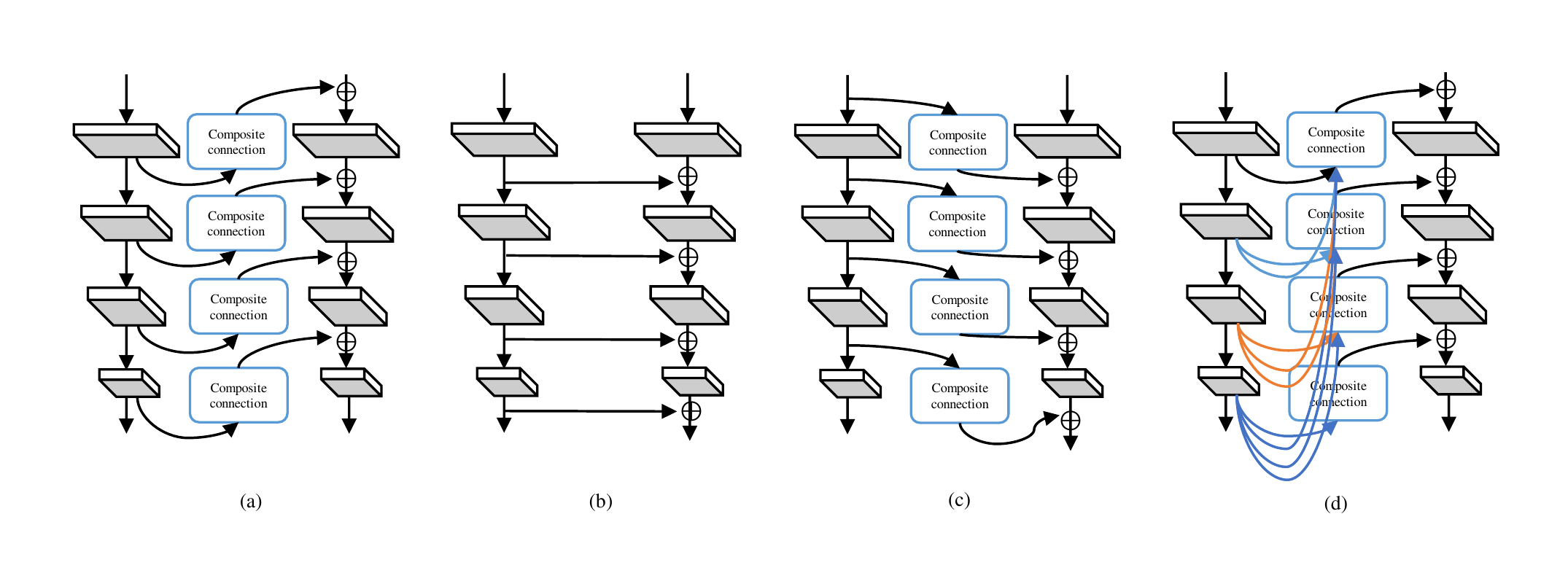}
	\caption{Four kinds of composite styles for Dual-Backbone architecture (an Assistant Backbone and a Lead Backbone). (a) Adjacent Higher-Level Composition (AHLC). (b) Same Level Composition (SLC). (c) Adjacent Lower-Level Composition (ALLC). (d) Dense Higher-Level Composition (DHLC). The composite connection denotes in blue boxes represents some simple operations, \textit{i.e.}, element-wise operation, scaling, 1$\times$1 Conv layer and bn layer.
	}
	\label{fig:three arch}
\end{figure*}

\subsection{Other possible composite styles}
\label{sec:opcab}

\subsubsection{Same Level Composition (SLC)} 
An intuitive and simple composite style is to fuse the output features from the same stage of backbones. This operation of Same Level Composite (SLC) can be formulated as:
\begin{equation}
x_k^l = F_k^l(x_k^{l-1} + x_{k-1}^{l-1}),  l\ge2.
\end{equation}
To be more specific, Figure~\ref{fig:three arch}.b illustrates the structure of SLC when $K=2$.

\subsubsection{Adjacent Lower-Level Composition (ALLC)} 
Contrary to AHLC, another intuitive composite style is to feed the output of the adjacent lower-level stage of the previous backbone to the succeeding backbone. This operation of Adjacent Lower-Level Composition (ALLC). The operation of Inverse Level Composite (ILC) can be formulated as:
\begin{equation}
x_k^l = F_k^l(x_k^{l-1} + g(x_{k-1}^{l+1})),  l\ge2.
\end{equation}
To be more specific, Figure~\ref{fig:three arch}.c illustrates the structure of ILC when $K=2$.

\subsubsection{Dense Higher-Level Composition (DHLC)} 
In DenseNet~\cite{huang2017densely}, each layer is connected to all subsequent layers to build a dense connection in a stage. Inspired by it, we can utilize dense composite connection in our CBNet architecture. The operation of DHLC can be expressed as follows:
\begin{equation}
x_k^l = F_k^l(x_k^{l-1} + \sum_{i=l}^L g_i(x_{k-1}^i)),  l\ge2.
\end{equation}
As shown in Figure~\ref{fig:three arch}.d, when $K=2$,
we assemble the features from all the higher-level stages in the Assistant Backbone, and add the composite features to the output features of the previous stage in the Lead Backbone.

\subsection{Architecture of detection network with CBNet }
\label{sec:det}
The CBNet architecture is applicable with various off-the-shelf object detectors without additional modifications to the network architectures.
In practice, we attach layers of the Lead Backbone with functional networks, RPN \cite{ren2015faster} , detection head \cite{Redmon_2016_CVPR,ren2015faster,zhang2018single,lin2017feature,he2017mask,cai18cascadercnn}.

\begin{table*}[t]
	\centering
	\small
	
	\begin{tabular}{c|ccc|ccc|ccc}
		\toprule
		Baseline detector  &Single &DB &TB  & $AP_{bbox}$ & $AP_{50}$ & $AP_{75}$  & $AP_{mask}$ & $AP_{50}$ & $AP_{75}$ \\
		\hline
		\multirow{3}{*}{ FPN + ResNet101}
		&\ding{51}&& &39.4 &61.5 &42.8 & - &- &- \\
		&&\ding{51}& &41.0&62.4&44.5& - &- &- \\
		&&&\ding{51} &\textbf{41.7}&\textbf{64.0}&\textbf{45.5}& - &- &- \\
		\hline
		\multirow{3}{*}{Mask R-CNN + ResNet101}
		&\ding{51}&& &40.0 &61.2 &43.6 & 35.9 & 57.9 & 38.0 \\
		&&\ding{51}& &41.8 &62.9 &45.6 & 37.0 & 59.5 & 39.3 \\
		&&&\ding{51} &\textbf{42.4}&\textbf{64.0}&\textbf{46.7} &\textbf{38.1}&\textbf{59.9}&\textbf{40.8}\\
		\hline
		\multirow{3}{*}{Cascade R-CNN + ResNet101}
		&\ding{51}&& &42.8 &62.1 &46.3& - &- &- \\
		&&\ding{51}& &44.3 &62.5 &48.1& - &- &- \\
		&&&\ding{51} &\textbf{44.9}&\textbf{63.9}&\textbf{48.9}& - &- &- \\
		
		\hline
		\multirow{3}{*}{Cascade Mask R-CNN + ResNeXt152}  &\ding{51}&&  & 48.3 & 67.0 & 52.8 & 41.0 &64.1 &44.2 \\
		&&\ding{51}& & 50.0 & 68.8 & 54.6 & 42.0 &64.6 &45.6 \\
		&&&\ding{51} & \textbf{50.7} & \textbf{69.8} & \textbf{55.5}& \textbf{43.3} &\textbf{66.9} &\textbf{46.8} \\
		
		\bottomrule
	\end{tabular}
	\caption{Detection results on the MS-COCO \texttt{test-dev} set. We report both object detection and instance segmentation results on four kinds of detectors to demonstrate the  effectiveness of CBNet. Single: with/without baseline backbone. DB: with/without Dual-Backbone architecture. TB: with/without Triple-Backbone architecture. Column 5-7 show the results of object detection while column 8-10 show the results of instance segmentation.}
	\label{table:detection result 1}
\end{table*}


\section{Experiments}
\label{section 4}
In this section, we present experimental results on the bounding box detection task and instance segmentation task of the challenging MS-COCO benchmark \cite{lin2014microsoft}. Following the protocol in MS-COCO, we use the \texttt{trainval35k} set for training, which is a union of 80k images from the train split and a random 35k subset of images from the 40k image validation split. We report COCO AP on the \texttt{test-dev} split for comparisons, which is tested on the evaluation server.


\subsection{Implementation details}

Baselines methods in this paper are reproduced by ourselves based on the Detectron framework~\cite{Detectron2018}. All the baselines are trained with single-scale strategy, except Cascade Mask R-CNN ResNeXt152. Specifically, the short side of input image is resized to 800, and the longer side is limited to 1333. We conduct experiments on a machine with 4 NVIDIA Titan X GPUs, CUDA 9.2 and cuDNN 7.1.4 for most experiments. In addition, we train Cascade Mask R-CNN with Dual-ResNeXt152 on a machine with 4 NVIDIA P40 GPUs and Cascade Mask R-CNN with Triple-ResNeXt152 on a machine with 4 NVIDIA V100 GPUs. The data augmentation is simply flipping the images. For most of the original baselines, batch size on a single GPU is two images. Due to the limitation of GPU memory for CBNet , we put one image on each GPU for training the detectors using CBNet. Meanwhile, we set the initial learning rate as the half of the default value and train for the same epoches as the original baselines. It is worth noting that, we do not change any other configuration of these baselines except the reduction of the initial learning rate and batch size.

During the inference, we completely use the configuration in the original baselines \cite{Detectron2018}. For Cascade Mask R-CNN with different backbones, we run both single-scale test and multi-scale test. And for other baseline detectors, we run single-scale test, in which the short side of input image is resized to 800, and the longer side is limited to 1333. It is noted that we do not utilize Soft-NMS \cite{bodla2017soft} during the inference for fair comparison .

\begin{table*}[t]
	\centering
	\small
	\begin{tabular}{l|c|ccc|ccc}
		\toprule
		Method & Backbone &$AP_{box}$ & $AP_{50}$ & $AP_{75}$ & $AP_{S}$ & $AP_{M}$ & $AP_{L}$  \\ 
		\hline
		\textit{one stage:} & & & & & & \\
		SSD512 \cite{liu2015ssd}&VGG16 &28.8 & 48.5 & 30.3 & 10.9 & 31.8 & 43.5 \\
		RetinaNet \cite{lin2017focal}&ResNeXt101&40.8 &61.1 & 44.1&24.1 &44.2 & 51.2\\
		RefineDet \cite{zhang2018single}*& ResNet101&41.8 & 62.9 & 45.7 & 25.6 & 45.1 & 54.1 \\
		CornerNet \cite{zhang2018single}*& Hourglass-104 &42.2&57.8&45.2&20.7&44.8&56.6 \\
		M2Det \cite{zhao2018m2det}*&VGG16&44.2&64.6&49.3&29.2&47.9&55.1\\
		FSAF \cite{zhu2019feature}*&ResNext-101&44.6&65.2&48.6&29.7&47.1&54.6\\
		NAS-FPN \cite{ghiasi2019fpn}&AmoebaNet&48.3&- &- & -&- &-\\
		\hline
		\textit{two stage:} & & & & & & \\
		Faster R-CNN \cite{ren2015faster}&VGG16 & 21.9 & 42.7&- &- &- &- \\
		R-FCN \cite{dai2016r}&ResNet101&29.9&51.9 &- &10.8 &32.8 & 45.0\\
		FPN \cite{lin2017feature}&ResNet101&36.2&59.1 &39.0 &18.2 &39.0 & 48.2\\
		Mask R-CNN \cite{he2017mask}&ResNet101&39.8 & 62.3 & 43.4 & 22.1 & 43.2& 51.2 \\
		Cascade R-CNN \cite{cai18cascadercnn}&ResNet101&42.8 &62.1&46.3&23.7&45.5&55.2  \\
		Libra R-CNN \cite{pang2019libra}&ResNext-101&43.0&64.0&47.0&25.3&45.6&54.6\\
		SNIP (model ensemble) \cite{singh2018analysis}*&- &48.3 & 69.7 & 53.7 & 31.4 & 51.6 & 60.7 \\
		SINPER \cite{singh2018sniper}*&ResNet101&47.6 &68.5 &53.4 &30.9 &50.6&60.7 \\
		Cascade Mask R-CNN \cite{Detectron2018}*&ResNeXt152 & 50.2 & 68.2 & 54.9 & 31.9 & 52.9 & 63.5\\
		MegDet (model ensemble) \cite{peng2018megdet}* & - & 52.5 & - & - & - & - & - \\
		\hline
		\textit{ours:(single model)} & & & & & & \\
		Cascade Mask R-CNN * &\textbf{\textit{Dual-ResNeXt152}}&\textbf{52.8} & \textbf{70.6} & \textbf{58.0} & \textbf{34.9} & \textbf{55.4} & \textbf{65.3}\\
		Cascade Mask R-CNN * &\textbf{\textit{Triple-ResNeXt152}}& \textbf{53.3} & \textbf{71.9} & \textbf{58.5} & \textbf{35.5} & \textbf{55.8} & \textbf{66.7}\\
		\bottomrule
	\end{tabular}
	\caption{Object detection comparison between our methods and state-of-the-art detectors on COCO \texttt{test-dev} set. * : utilizing multi-scale testing.} 
	\label{table:detection compare} 
\end{table*}


\subsection{Detection results}

To demonstrate the effectiveness of the proposed CBNet, we conduct a series of experiments with the baselines of state-of-the-art detectors, \textit{i.e.}, FPN \cite{lin2017feature}, Mask R-CNN \cite{he2017mask} and Cascade R-CNN \cite{cai18cascadercnn}, and the results are reported in Table \ref{table:detection result 1}. In each row of Table~\ref{table:detection result 1}, we compare a baseline (provided by Detectron \cite{Detectron2018}) with its variants using the proposed CBNet, and one can see that our CBNet consistently improves all of these baselines with a significant margin. More specifically, the mAPs of these baselines increase by 1.5 to 3 percent.

Furthermore, as presented in Table \ref{table:detection compare}, a new state-of-the-art detection result of 53.3 mAP on the MS-COCO benchmark is achieved by Cascade Mask R-CNN baseline equipped with the proposed CBNet. Notably, this result is achieved just by single model, without any other improvement for the baseline besides taking CBNet as backbone. Hence, this result demonstrates great effectiveness of the proposed CBNet architecture. 


Moreover, as shown in Table \ref{table:detection result 1}, the proposed CBNet also improves the performances of the baselines for instance segmentation. Compared with bounding boxes prediction (\textit{i.e.}, object detection), pixel-wise classification (\textit{i.e.}, instance segmentation) tends to be more difficult and requires more representational features. And these results demonstrate the effectiveness of CBNet again.

\subsection{Comparisons of different composite styles}

We further conduct experiments to compare the suggested composite style AHLC with other possible composite styles illustrated in Figure \ref{fig:three arch}, including SLC, ALLC, and DHLC. All of these experiments are conducted based on the Dual-Backbone architecture and the baseline of FPN ResNet101.

\textbf{SLC \textit{v.s.} AHLC}
As presented in Table \ref{table:detection result 2}, SLC gets even worse result than the original baseline. We think the major reason is that the architecture of SLC will bring serious parameter redundancy. To be more specific, the features extracted by the same stage of the two backbones in CBNet are similar, hence SLC cannot learn more semantic information than using single backbone. In other words, the network parameters are not fully utilized, but bring much difficulty on training, leading to a worse result.

\textbf{ALLC \textit{v.s.} AHLC}
As shown in Table \ref{table:detection result 2}, there is a great gap between ALLC and AHLC. We infer that, in our CBNet, if we directly add the lower-level (i.e., shallower) features of the previous backbone to the higher-level (i.e., deeper) ones of the succeeding backbone, the semantic information of the latter ones will be largely harmed. On the contrary, if we add the deeper features of the previous backbone to the shallow ones of the succeeding backbone, the semantic information of the latter ones can be largely enhanced.

\textbf{DHLC \textit{v.s.} AHLC}
The results in Table \ref{table:detection result 2} show that DHLC does not bring performance improvement as AHLC, although it adds more composite connections than AHLC. We infer that, the success of Composite Backbone Network lies mainly in the composite connections between adjacent stages, while the other composite connections do not enrich much feature since they are too far away. 

Obviously, CBNets of these composite styles have same amount of the network parameter (i.e., about twice amount of the network parameters than single backbone), but only AHLC brings optimal detection performance improvement. These experiment results prove that \textit{only increasing parameters or adding additional backbone may not bring better result.} Moreover, these experiment also show that \textit{composite connections should be added properly.} Hence, these experiment results actually demonstrate that \textit{the suggested composite style AHLC is effective and nontrivial.}

\begin{table}[H]
	\small
	\centering 
	\begin{tabular}{c | c| c c c } 
		\toprule 
		DB &Composite style & $AP_{box}$ & $AP_{50}$ & $AP_{75}$  \\ 
		\hline	
		& - & 39.4 & 61.5 & 42.8\\
		\ding{51} &SLC & 38.9 & 60.8 & 42.0 \\
		\ding{51} &ALLC & 36.5 & 57.6 & 39.6 \\
		\ding{51} &ADLC & 40.7 & 61.8 & 44.0 \\
		\ding{51}&AHLC	& \textbf{41.0} & \textbf{62.4} & \textbf{44.5}\\
		
		\bottomrule 
	\end{tabular}
	\caption{Comparison between different composite styles, the baseline is FPN ResNet101 \cite{lin2017feature}. DB: with/without Dual-Backbone. "SLC" represents Same Level Composition, "ALLC" represents Adajacent Lower-Level Composition,"ADLC" is Adjacent Dense Level Composition and "AHLC" is Adjacent Higher-Level Composition. } 
	\label{table:detection result 2} 
\end{table}


\subsection{Sharing weights for CBNet}
Due to using more backbones, CBNet brings the sharp increase of network parameters. To further demonstrate that the improvement of detection performance mainly comes from the composite architecture rather than the increase of network parameters, we conduct experiments on FPN, with the configuration of sharing the weighs of two backbones in Dual-ResNet101, and the results are shown in Table \ref{table:share weighs}. We can see that when sharing the weights of backbones in CBNet, the increment of parameters is negligible, but the detection result is still much better than the baseline (\textit{e.g.}, mAP 40.4 \textit{v.s.} 39.4). However, when we do not share the weights, the improvement is not so much (mAP from 40.4 to 41.0), which proves that \textit{it is the composite architecture that boosts the performance dominantly, rather than the increase of network parameters}.

%


\begin{table}[H]
	\small
	\centering 
	\begin{tabular}{c|c|c|l|c} 
		\toprule 
		Baseline detector &\textit{DB} & Share  & $AP_{box}$ &  \textit{mb} \\ 
		\hline
		\multirow{3}{*}{FPN + ResNet101} & & & 39.4 & 470\\
		& \ding{51} & \ding{51}  &40.4& 492\\
		& \ding{51}  &  & 41.0 & 815\\
		\bottomrule 
	\end{tabular}
	
	\caption{Comparison of with/without sharing weights for Dual-Backbone architecture. DB: with/without Dual-Backbone. Share: with/without sharing weights. $AP_{box}$: detection results on COCO \texttt{test-dev} dataset. \textit{mb}: the model size.} 
	\label{table:share weighs} 
\end{table}

\subsection{Number of backbones in CBNet }
\begin{figure}[H]
	\centering
	\includegraphics[scale=0.22]{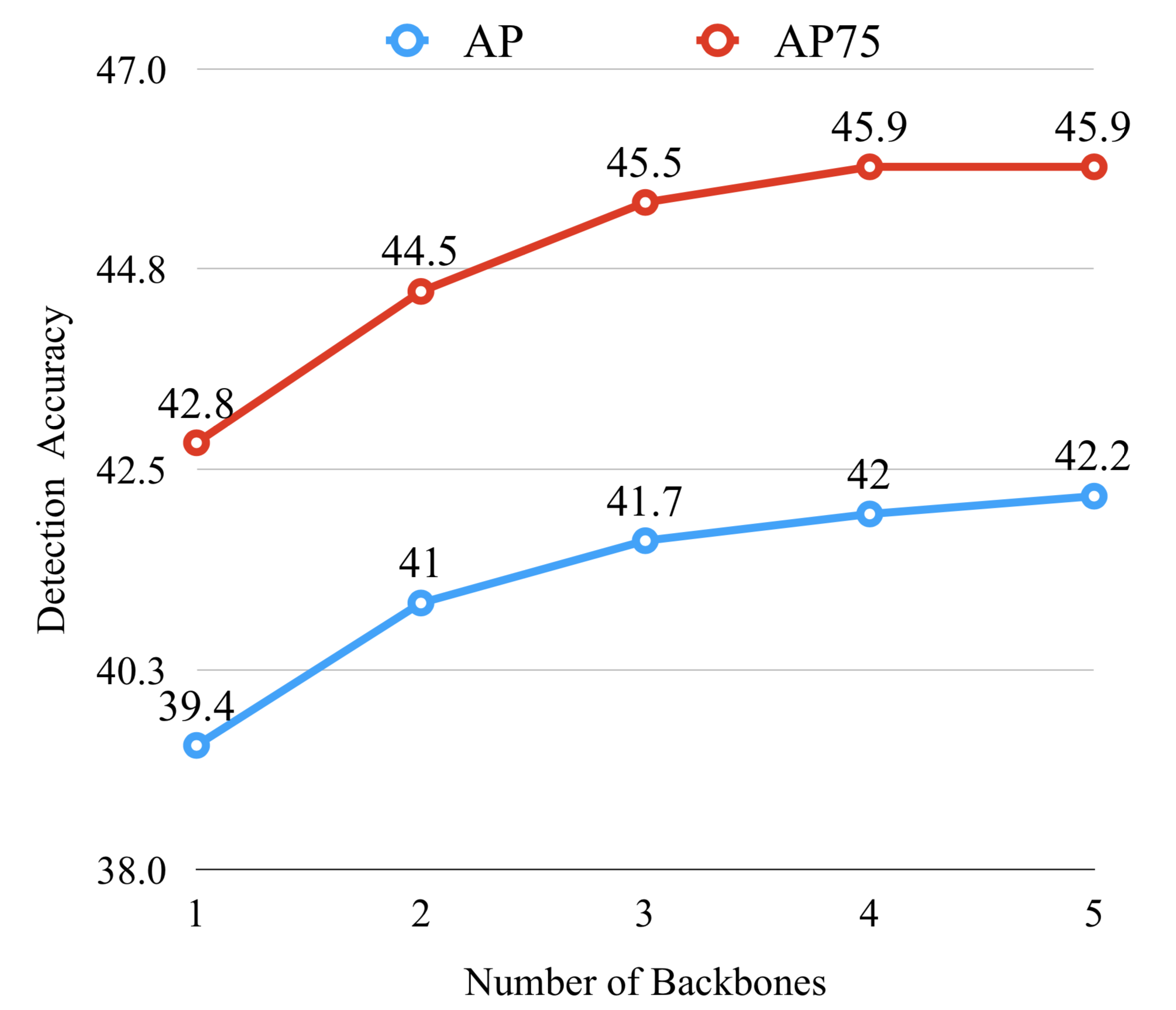}
	\caption{Object detection results on the MS-COCO \texttt{test-dev} dataset using different number of backbones in CBNet architecture based on FPN ResNet101. }
	\label{fig:more}
\end{figure}

We conduct experiments to investigate the relationship between the number of backbones in CBNet and the detection performance by taking FPN-ResNet101 as the baseline, and the results are shown in Figure~\ref{fig:more}. It can be noted that the detection mAP steadily increases with the number of backbones, and tends to converge when the number of backbones reaches three. Hence, considering the speed and memory cost, we suggest to utilize Dual-Backbone and Triple-Backbone architectures.

\begin{figure}[H]
	\centering
	\includegraphics[scale=0.25]{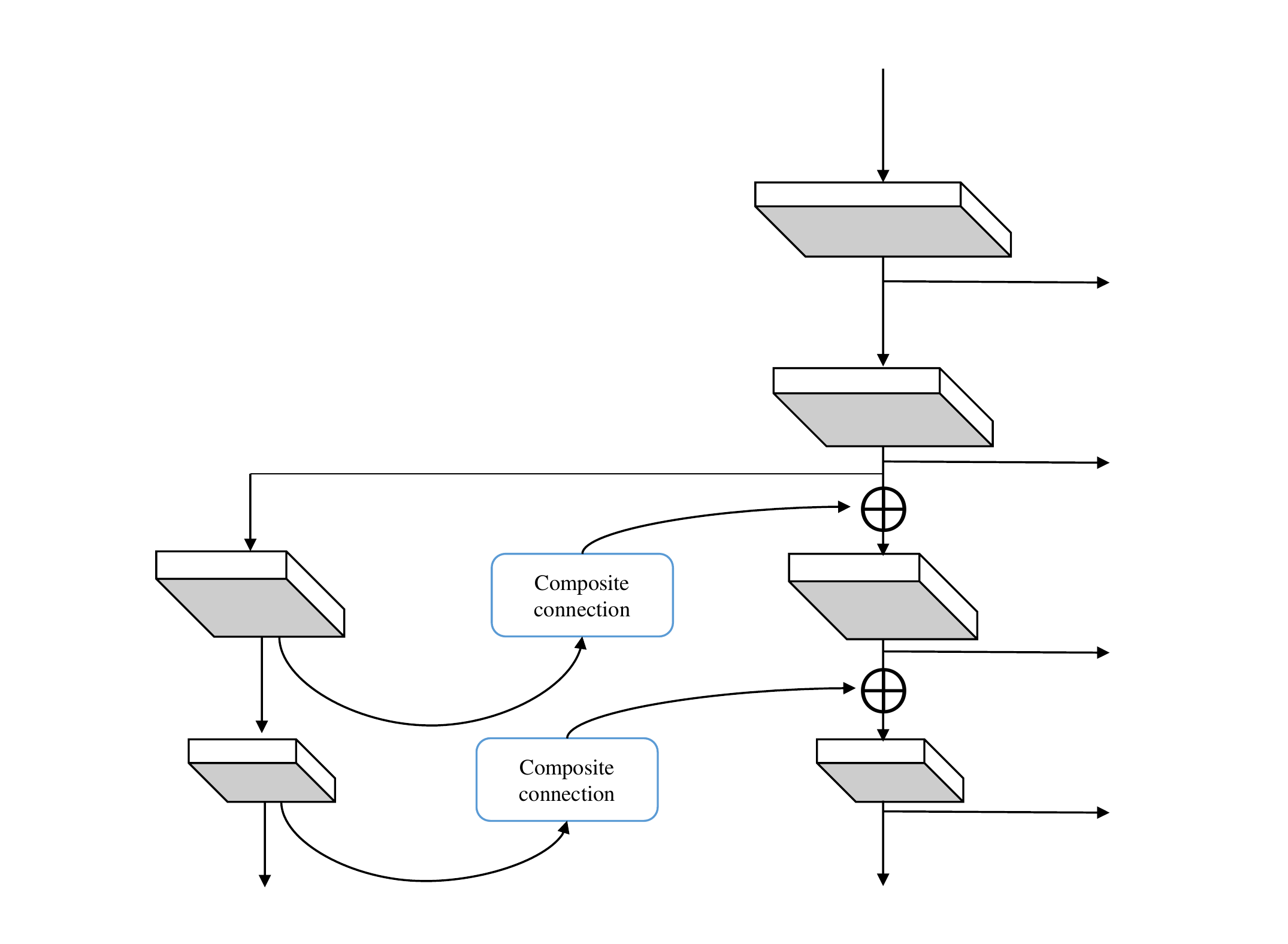}
	\caption{An accelerated version of CBNet ($K=2$).}
	\label{fig:db_jianzhi}
\end{figure}

\begin{table}[H]
	\centering
	\small
	\begin{tabular}{c|l|c|c|c}
		\toprule
		Baseline detector & DB & $\Psi$ & $AP_{box}$  & \textit{fps} \\ 
		\hline
		\multirow{3}{*}{FPN + ResNet101} &  &  & 39.4 & 8.1\\
		& \ding{51} & &\textbf{41.0} & 5.5 \\
		& \ding{51} & \ding{51} & \textbf{40.8} & 6.9 \\
		\bottomrule
	\end{tabular}
	\caption{Performance comparison between the original DB and the accelerated vesion. DB: with/without Dual-Backbone. $\Psi$: with/without the acceleration modification of the CBNet architecture illustrated in Figure \ref{fig:db_jianzhi}.} 
	\label{table:speed issue} 
\end{table}

\subsection{An accelerated version of CBNet
}

The major drawback of the proposed CBNet is that it will slows down the inference speed of the baseline detector since it uses more backbones to extract features thus increases the computation complexity. For example, as shown in Table \ref{table:speed issue}, DB increases the AP of FPN by 1.6 percent but slows down the detection speed from 8.1 fps to 5.5 fps. To alleviate this problem, we further propose an accelerated version of the CBNet as illustrated in Figure \ref{fig:db_jianzhi}, by removing the two early stages of the Assistant Backbone. As demonstrated in Table \ref{table:speed issue}, this accelerated version can significantly improve the speed (from 5.5 fps to 6.9 fps) while not harming the detection accuracy (i.e., AP) a lot (from 41.0 to 40.8).


\subsection{Effectiveness of basic feature enhancement by CBNet}

We think the root cause that our CBNet can performs much better than the single backbone network for object detection task is: \textit{it can extract more representational basic features than the original single backbone network which is originally designed for classification problem.} To verify this, as illustrated in Figure 6, we visualize and compare the intermediate the feature maps extracted by our CBNet and the original single backbone in the detectors for some examples. The example image in Figure 6 contains two foreground objects: a person and a tennis ball. Obviously, the person is the large-size object and the tennis ball is the small-size object. Hence, we correspondingly visualize the large scale feature maps (for detecting small objects) and the small scale feature maps (for detecting large objects) extracted by our CBNet and the original single backbone. One can see that, the feature maps extracted by our CBNet consistently have stronger activation values at the foreground object and weaker activation values at the background. This visualization example shows that our CBNet is more effective to extract representational basic features for object detection.

\begin{figure}[th]
	\centering
	\includegraphics[scale=1.1]{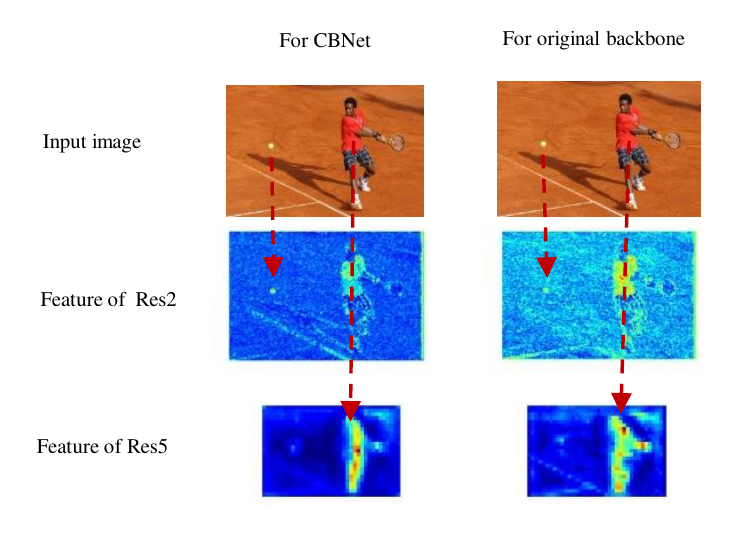}
	\caption{Visualization comparison of the features extracted by our CBNet (Dual-ResNet101) and the original backbone (ResNet101). The baseline detector is FPN-ResNet101. For each backbone, we visualize the Res2 and Res5 according to the size of the foreground objects, by averaging feature maps along channel dimension. It is noteworthy that feature maps come from the CBNet are more representational since they have stronger activation values at the foreground object and weaker activation values at the background. \textbf{Best view in color}.}
	\label{fig:four}
\end{figure}

\section{Conclusion}
\label{section 6}
In this paper, a novel network architecture called \emph{Composite Backbone Network} (CBNet) is proposed to boost the performance of state-of-the-art object detectors. CBNet consists of a series of backbones with same network structure and uses composite connections to link these backbones. Specifically, the output of each stage in a previous backbone flows to the parallel stage of the succeeding backbone as part of inputs through composite connections. Finally, the feature maps of the last backbone namely Lead Backbone are used for object detection.
Extensive experimental results demonstrate that the proposed CBNet is beneficial for many state-of-the-art detectors, such as FPN, Mask R-CNN, and Cascade R-CNN, to improve their detection accuracy. To be more specific, the mAPs of the detectors mentioned above on the COCO dataset are increased by about 1.5 to 3 percent, and a new state-of-the art result on COCO with the mAP of 53.3 is achieved by simply integrating CBNet into the Cascade Mask R-CNN baseline. Simultaneously, experimental results show that it is also very effective to improve the instance segmentation performance. Additional ablation studies further demonstrate the effectiveness of the proposed architecture and the composite connection module.

\bibliographystyle{aaai} 
\bibliography{egbib}
\end{document}